\documentclass{article}
\usepackage{CJKutf8}
\usepackage{ijcai25}
\usepackage{multirow}
\usepackage{color,xcolor}
\usepackage{bbding}
\usepackage{times}
\usepackage{soul}
\usepackage{url}
\usepackage[hidelinks]{hyperref}
\usepackage[utf8]{inputenc}
\usepackage[small]{caption}
\usepackage{graphicx}
\usepackage{amsmath}
\usepackage{amsthm}
\usepackage{booktabs}
\usepackage{algorithm}
\usepackage{algorithmic}
\usepackage[switch]{lineno}

\title{Chinese Spelling Correction: A Comprehensive Survey of \\Progress, Challenges, and Opportunities}

\author{
Changchun Liu$^1$, Kai Zhang$^1$, Junzhe Jiang$^1$,\\ Zixiao Kong$^1$, Qi Liu$^1$, Enhong Chen$^1$
\affiliations
$^1$State Key Laboratory of Cognitive Intelligence, University of Science and Technology of China
\emails
\{changchun\_liu, jzjiang, kzix\}@mail.ustc.edu.cn,
\{kkzhang08, qiliuql, cheneh\}@ustc.edu.cn
}

\begin{document}

\maketitle

\begin{abstract}
Chinese Spelling Correction (CSC) is a critical task in natural language processing, aimed at detecting and correcting spelling errors in Chinese text. This survey provides a comprehensive overview of CSC, tracing its evolution from pre-trained language models to large language models, and critically analyzing their respective strengths and weaknesses in this domain. Moreover, we further present a detailed examination of existing benchmark datasets, highlighting their inherent challenges and limitations. Finally, we propose promising future research directions, particularly focusing on leveraging the potential of LLMs and their reasoning capabilities for improved CSC performance. \textbf{\emph{To the best of our knowledge, this is the first comprehensive survey dedicated to the field of CSC}}. We believe this work will serve as a valuable resource for researchers, fostering a deeper understanding of the field and inspiring future advancements.

\end{abstract}

\section{Introduction}
Chinese Spelling Correction (CSC) is a critical research area in Natural Language Processing (NLP), dedicated to identifying and correcting spelling errors in Chinese text. The recent explosion of user-generated content across social networks, e-commerce platforms, and online education has led to a corresponding surge in spelling errors, which can significantly degrade the performance of downstream NLP applications. Addressing these errors is crucial for enhancing the accuracy of information retrieval and recommendation systems in domains like social media and online shopping, as well as facilitating automated assessment and personalized learning in education. Therefore, advancements in CSC technology are of paramount importance. 

The primary sources of errors in CSC typically stem from users' input method editors (IMEs)~\cite{hu-etal-2024-cscd}, as well as inaccuracies introduced by automatic speech recognition (ASR) and optical character recognition (OCR) systems. Based on the characteristics of these error sources, the IME and ASR originated errors are mainly related to the similar pronunciation (or pinyin) of Chinese characters, while OCR originated errors result primarily from similarities in the shapes of characters. In addition, statistical analyses reveal that 83\% of errors are due to similar pronunciations, while 48\% are attributed to similar shapes~\cite{liu2010visually}.

Early research in CSC relied heavily on rule-based and statistical methods. These methods employed predefined rules, large-scale corpora, and language models such as N-grams to rectify errors through statistical analysis. Subsequently, non-autoregressive pre-trained language models like BERT have emerged as the dominant paradigm in CSC, leading to a proliferation of related studies. Some research efforts~\cite{li-etal-2022-improving-chinese,liang-etal-2023-disentangled} have focused on effectively extracting and utilizing the phonetic and visual features of Chinese characters through various methodologies, while striving to accurately interpret semantics within erroneous contexts. In parallel, the traditional detector-corrector framework~\cite{huang-etal-2023-frustratingly,wu-etal-2024-bi} has been continually optimized and refined to get better performance.

Recent advancements in large language models (LLMs), such as GPT-4o and DeepSeek, have garnered significant attention due to their remarkable semantic understanding and reasoning capabilities. However, only a limited number of studies have explored the integration of LLMs in traditional models~\cite{liu-etal-2024-arm} or fine-tuned LLMs~\cite{li-etal-2024-c} for the CSC. This is mainly because LLMs, as autoregressive models, face fatal challenges in CSC tasks. Issues such as variable output length and tendencies toward overcorrection significantly impact their effectiveness. Consequently, the application of LLMs to the CSC tasks remains a largely untapped area, offering considerable potential for future research and development.

Despite the extensive research conducted on CSC, a systematic review summarizing the findings, providing a comprehensive overview of the field, and identifying future research directions is currently lacking. Therefore, this survey aims to provide readers a thorough understanding of the fundamental concepts of the CSC task, current research trends, existing challenges, and promising future directions.

To facilitate understanding of CSC techniques for readers from diverse backgrounds, this survey is organized as follows: \textbf{\emph{in Section~\ref{sec2}}}, we introduce the formal definition of the CSC task and its key characteristics. \textbf{\emph{Section~\ref{sec3} and~\ref{sec4}}} provide an in-depth discussion of the research methods applied to CSC across various stages of development, with a focus on model structure design and the extraction of Chinese character information and semantics under PLMs. \textbf{\emph{In Section~\ref{sec5}}}, we review existing datasets and evaluation criteria for model performance in CSC. \textbf{\emph{Section~\ref{sec6}}} discusses current CSC challenges in PLMs, LLMs, and existing datasets. Besides, we propose future research directions to address these challenges. Finally, \textbf{\emph{in Section~\ref{sec7}}}, we summarize this survey.

\begin{figure*}[t]
  \includegraphics[width=\textwidth]{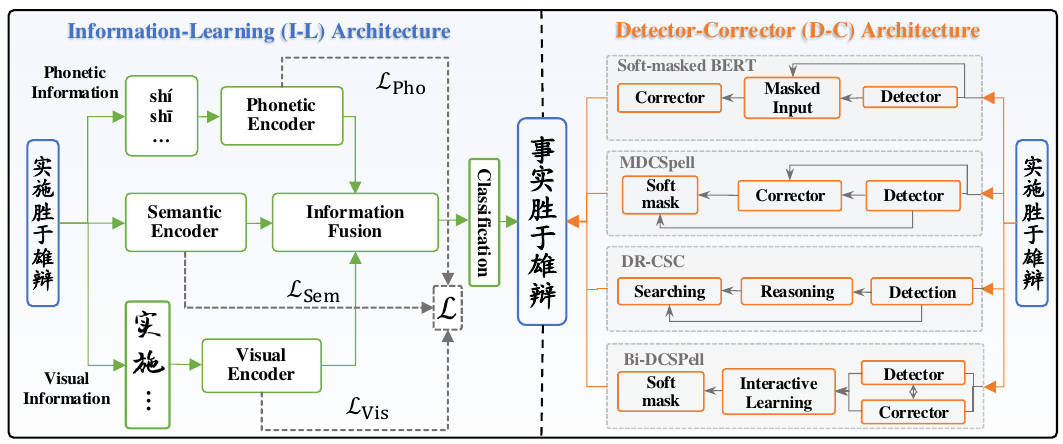}
  \caption{Two classic architectures of CSC model are presented. On the left, the I-L architecture illustrates the information flow, where some models integrate Chinese character information into the model's training and reasoning, while others incorporate it solely into the loss function. On the right, the D-C architecture demonstrates the information flow of four classic D-C models. The similarities and differences between them are clearly depicted. The English translation of the modified sentence ``\begin{CJK}{UTF8}{gkai}事实胜于雄辩\end{CJK}'' is ``Facts speak louder than words''.}
  \label{fig1}
\end{figure*}

\section{Definition of the CSC}\label{sec2}
\subsection{Formal Definition}
Chinese Spelling Correction (CSC) can be formally defined as follows: Given a Chinese sentence $\boldsymbol{X}=\{x_1,x_2,\ldots,x_n\}$ of $n$ characters that may include erroneous characters. We use $\boldsymbol{Y}=\{y_1,y_2,\ldots,y_n\}$ to represent the corresponding correct sentence. The sentence $\boldsymbol{X}$ and $\boldsymbol{Y}$ have the same length. The objective of the CSC is to detect and correct the erroneous characters by generating a prediction $\boldsymbol{\hat{Y}}=\{\hat{y_1},\hat{y_2},\ldots,\hat{y_n}\}$ for the input $\boldsymbol{X}$, where $\hat{y_i}$ is the character predicted for $x_i$. The primary mission of the CSC lies in accurately detecting the erroneous characters and predicting their correct counterparts in $\boldsymbol{Y}$.

\subsection{Characteristics of the CSC Tasks}
In this section, we provide a detailed description of the key features of the CSC. By dividing ``Chinese Spelling Correction" into \textbf{\emph{``Chinese"}}, \textbf{\emph{``Spelling"}}, and \textbf{\emph{``Correction"}}, we can clearly understand its essential characteristics.

The \textbf{\emph{``Chinese"}} represents the language used in the CSC task is Chinese. Chinese is an ideographic language, where the shape of the characters is closely tied to their meaning. In contrast, English is a phonetic language, where letters represent distinct phonemes that are combined into words to convey meaning through pronunciation. Moreover, the grammatical structure of them is quite different. For instance, unlike English, Chinese does not employ spaces to divide words. Instead, words must be segmented according to context. As a result, despite numerous studies on correcting errors in English, these cannot be directly utilized for CSC, necessitating a model tailored specifically for Chinese.

The \textbf{\emph{``Spelling"}} means that the CSC focuses on characters themselves rather than the overall text. The goal is to correct the misuse of individual Chinese character, rather than adjusting the entire sentence structure. It is worth noting that this does not imply a disregard for the overall meaning, but emphasizes the modification focusing on character errors. ``Spelling" also pointed out that the error in the CSC study was an alignment error. That is, the sentence length remains unchanged before and after modification, with only ``replacement" being used as a modification method, rather than ``addition" or ``deletion". From this perspective, CSC can be viewed as a sequence labeling task.

The \textbf{\emph{``Correction"}} points directly to the primary target of the CSC: identifying and correcting errors in the sentence.
\section{Methods Based on PLMs}\label{sec3}
\begin{table*}[ht]
    \small
    \centering
    \begin{tabular}{c|c|c|c|c|c|c|c}
       \toprule
        \multirow{2}{*}{Research} & \multirow{2}{*}{Model Name} & \multirow{2}{*}{Structure} & \multicolumn{2}{c|}{Phonetic Information} & \multicolumn{2}{c|}{Visual Information}& \multirow{2}{*}{Con-Set} \\ \cmidrule(lr){4-7}
        & & & Emb  & Spe-Loss & Emb  & Spe-Loss \\ \midrule
        \cite{hong-etal-2019-faspell}& FASPell&I-L&\Checkmark &\XSolidBrush  &\Checkmark &\XSolidBrush &\XSolidBrush  \\ \cmidrule(lr){1-8}
        \cite{cheng2020spellgcn}& SpellGCN&I-L&\Checkmark & \XSolidBrush &\Checkmark& \XSolidBrush&\Checkmark \\ \cmidrule(lr){1-8}
        \cite{zhang-etal-2020-spelling}& Soft-Masked BERT&D-C&\XSolidBrush &  \XSolidBrush&\XSolidBrush & \XSolidBrush & \Checkmark\\ \cmidrule(lr){1-8}
        \cite{liu2021plome}& PLOME&I-L&\Checkmark & \Checkmark& \Checkmark&\XSolidBrush &\Checkmark  \\ \cmidrule(lr){1-8}
        \cite{huang2021phmospell}&PHMOSpell&I-L&\Checkmark &\XSolidBrush &\Checkmark &\XSolidBrush &\XSolidBrush  \\ \cmidrule(lr){1-8}
        \cite{guo2021global}&GAD&I-L& \XSolidBrush& \XSolidBrush&\XSolidBrush &\XSolidBrush &\Checkmark  \\ \cmidrule(lr){1-8}
        \cite{zhang2021correcting}&MLM-phonetics &D-C& \Checkmark& \Checkmark& \XSolidBrush& \XSolidBrush& \Checkmark \\ \cmidrule(lr){1-8}
        \cite{wang2021dynamic}& DCN&I-L&\Checkmark &\XSolidBrush &\XSolidBrush & \XSolidBrush&\Checkmark  \\ \cmidrule(lr){1-8}
        \cite{xu-etal-2021-read}&REALISE &I-L&\Checkmark & \Checkmark& \Checkmark& \Checkmark&\XSolidBrush  \\ \cmidrule(lr){1-8}
        \cite{ji-etal-2021-spellbert}&SpellBERT &I-L&\Checkmark &\Checkmark &\Checkmark &\Checkmark &\Checkmark  \\ \cmidrule(lr){1-8}
        \cite{zhu2022mdcspell}&MDCSpell &D-C& \XSolidBrush&\XSolidBrush & \XSolidBrush&\XSolidBrush &\XSolidBrush  \\ \cmidrule(lr){1-8}
        \cite{shulin2022craspell}&CRASpell &I-L&\XSolidBrush &\XSolidBrush & \XSolidBrush&\XSolidBrush & \Checkmark \\ \cmidrule(lr){1-8}
        \cite{li-etal-2022-improving-chinese}& SCOPE&I-L& \XSolidBrush&\Checkmark & \XSolidBrush&\XSolidBrush & \Checkmark \\ \cmidrule(lr){1-8}
        \cite{li-etal-2022-wspeller}&WSpeller &I-L&\XSolidBrush &\XSolidBrush &\XSolidBrush & \XSolidBrush&\XSolidBrush  \\ \cmidrule(lr){1-8}
        \cite{li-etal-2022-learning-dictionary}&LEAD &I-L&\XSolidBrush&\Checkmark &\XSolidBrush&\Checkmark &\Checkmark  \\ \cmidrule(lr){1-8}
        \cite{wei-etal-2023-ptcspell}&PTCSpell&I-L&\Checkmark &\Checkmark & \Checkmark&\Checkmark &\Checkmark \\ \cmidrule(lr){1-8}
        \cite{liang-etal-2023-disentangled}&DORM&I-L&\Checkmark&\Checkmark&\XSolidBrush&\XSolidBrush&\Checkmark \\ \cmidrule(lr){1-8}
        \cite{huang-etal-2023-frustratingly}&DR-CSC&D-C&\XSolidBrush &\XSolidBrush & \XSolidBrush&\XSolidBrush & \Checkmark \\ \cmidrule(lr){1-8}
        \cite{wu-etal-2024-bi}&Bi-DCSpell&D-C&\XSolidBrush&\XSolidBrush&\XSolidBrush&\XSolidBrush&\XSolidBrush \\ \cmidrule(lr){1-8}
        \cite{yin-etal-2024-error}&RERIC&I-L&\Checkmark & \XSolidBrush& \Checkmark& \XSolidBrush&\XSolidBrush  \\ \cmidrule(lr){1-8}
        \cite{lin-etal-2024-modalities}&UGMSC&I-L&\Checkmark&\XSolidBrush&\Checkmark&\XSolidBrush&\Checkmark \\ 
        \bottomrule
        
    \end{tabular}
    \caption{The model's architecture classification and the utility pattern of Chinese character phonetic and visual information. ``Emb'' indicates whether the information is embedded within the model, ``Spe-Los'' specifies whether a specialized loss function is designed for the information,  ``Con-Set'' denotes whether a confusion set is used, ``I-L'' and ``D-C''indicates Information-Learning and Detector-Corrector architecture.}
    \label{tab1}
\end{table*}
During earlier periods, CSC research was relatively fragmented and limited. With the advancement of deep learning, the field of CSC has entered a revolutionary phase marked by rapid progress, with the most important achievements so far being based on PLMs. During this phase, substantial research efforts have focused on leveraging PLMs to extract semantic information from sentences as well as the phonetic and visual features of Chinese characters, optimizing model architectures and enhancing pretraining or finetuning methods.

Although numerous studies have been conducted, most model structures can be categorized into two main types, models based on Information-Learning architecture (I-L) and models based on Detector-Corrector architecture (D-C), as illustrated in \textbf{Figure~\ref{fig1}}. These two categories are not entirely independent, as some studies exhibit overlaps. In addition, for I-L architecture, various methods are also being developed to learn Chinese character information. \textbf{Table~\ref{tab1}} summarizes recent model architectures and their approaches to learning Chinese character information. 

This section is divided into two parts for a detailed discussion. The first part examines how models learn character information and integrate them with sentence semantics, which constitutes the core of the I-L architecture. The second part introduces the development process of D-C architecture. 
\subsection{Learning and Utilization of Character Information}
Since most sources of errors involve similar Chinese characters, it is essential to learn phonetic and visual information from characters and evaluate their similarity. Therefore, this part will introduce in detail how to learn and use the phonetic and visual information of Chinese characters.
\subsubsection{I. Confusion Set} 
 An effective approach to identify similar characters is to use a confusion set. The confusion set is a collection of characters that are easily confused, typically constructed based on similarities in phonetics and visuals. Despite its limitations, such as restricted coverage, lack of adaptability, and potential biases, confusion sets are still used in studies~\cite{li-etal-2022-improving-chinese,huang-etal-2023-frustratingly}, such as during the pretraining phase, as once constructed, they eliminate the need for additional computations to determine character similarity.
\subsubsection{II. Learning of Phonetic Information} 
Most models utilize the pinyin sequences of Chinese characters to represent their phonetic information, with only a few exceptions, such as Tacotron2~\cite{huang2021phmospell}. Pinyin effectively captures the phonetic characteristics in a simple and accurate manner. While some models~\cite{liang-etal-2023-disentangled} process the pinyin sequence as a whole, others adopt a more granular approach~\cite{li-etal-2022-improving-chinese} by distinguishing between initials, finals, and tones.
\subsubsection{III. Learning of Visual Information} 
There are various approaches to obtaining visual information of Chinese characters. Some studies~\cite{cheng2020spellgcn} use the strokes of Chinese characters directly as visual information. However, the same stroke sequence can form entirely different characters when arranged in different structural combinations. To address this, some research~\cite{li-etal-2022-wspeller} incorporates an ideographic description sequence comprising structural components and strokes to obtain the tree structure of Chinese characters. In the deep learning era, models like ResNet and VGG19 have been explored to extract the visual features of Chinese characters~\cite{xu-etal-2021-read,huang2021phmospell}. However, statistical analysis and error sources indicate that errors caused by visual similarity are relatively rare. As a result, an increasing number of models have opted to exclude using visual information.
\subsubsection{IV. Utilization of Character Information} 
After extracting those information, utilizing it effectively is essential, which can be achieved through three primary approaches described in detail below.

\textbf{Adjusting the masking strategy}~\cite{zhang2021correcting,li-etal-2022-wspeller}. Unlike using the random replacement or the label ``[MASK]'' in the pretraining process of BERT-like models, characters are instead randomly replaced with others that share similar phonetics or visual features. These substitutions are sourced from a confusion set or determined by using similarity-based algorithms like Levenshtein Distance. This strategy enables the model to learn the relationships between phonetically or visually similar characters, thereby prioritizing the replacement of similar characters in potentially erroneous positions. Additionally, to mitigate overfitting, a certain proportion of characters is randomly replaced or masked during pretraining.

\textbf{Specialized loss function}. Some studies integrate Chinese character information into the loss function, allowing the model to learn relevant features during training. LEAD~\cite{li-etal-2022-learning-dictionary} integrates phonetic, shape, and meaning information into the contrastive function by creating positive and negative samples, and these samples optimize the CSC model within a unified contrastive learning framework. SCOPE~\cite{li-etal-2022-improving-chinese} incorporates two parallel decoders: one for CSC and the other for auxiliary Chinese pronunciation prediction. It optimizes performance by dynamically balancing the primary task CSC and the auxiliary task, using an adaptive weight adjustment mechanism that places greater emphasis on mistakes with similarly sounding characters.

\textbf{Embedded in models}. Some studies integrate Chinese character information into various parts of the model, such as word embedding or classification part. Specifically, SpellGCN~\cite{cheng2020spellgcn} integrates the phonetic and shape similarities of Chinese characters using a GCN. It constructs phonetic and shape similarity graphs and propagates features through the GCN, enabling similar Chinese characters to influence one another. In addition, PHMOSpell~\cite{huang2021phmospell} and REALISE~\cite{xu-etal-2021-read} leverage multimodal information to enhance CSC by integrating phonetic, visual, and semantic information. PHMOSpell employs text-to-speech tasks to learn latent representations of Chinese character pronunciation and utilizes VGG19 to extract glyph features. Additionally, it incorporates an adaptive gating mechanism to dynamically regulate the fusion of phonetic, visual, and semantic information. REALISE employs a hierarchical pinyin encoder to extract phonetic features at both the character and sentence levels while utilizing ResNet to extract glyph features and uses a selective modality fusion mechanism, which dynamically adjusts the fusion ratio of semantic, pinyin, and glyph information through gating units.
\begin{table*}[ht]
    \small
    \centering
    \begin{tabular}{l|l|l|c|c|l|c|c|c}
    \toprule
    \multicolumn{1}{c|}{Research} & \multicolumn{1}{c|}{Name} & \multicolumn{1}{c|}{TNS} & ASL & AEPS & \multicolumn{1}{c|}{Source} & Real Error & Expert & Split \\ \midrule
    ~\cite{wu2013chinese}&SIGHAN13&1700  &60.8&0.92 & Chinese Beginners& \Checkmark&\Checkmark&\Checkmark \\ \cmidrule(lr){1-9}
    ~\cite{yu2014overview}& SIGHAN14& 4499& 49.5& 1.01& Chinese Beginners& \Checkmark&\Checkmark&\Checkmark \\ \cmidrule(lr){1-9}
    ~\cite{tseng2015introduction}&SIGHAN15 & 3439&31.0 &0.95& Chinese Beginners& \Checkmark&\Checkmark&\Checkmark \\ \cmidrule(lr){1-9}
    ~\cite{wang-etal-2018-hybrid}&Wang271K &271329 &42.6 & 1.40&Official Newspapers &\XSolidBrush &\XSolidBrush &\XSolidBrush \\ \cmidrule(lr){1-9}
    ~\cite{jiang2022mcscset}& MCSC&196495 &10.9 &0.93 &A Medical Website &\Checkmark & \Checkmark& \Checkmark\\ \cmidrule(lr){1-9}
    ~\cite{lv2023general}&ECSPell & 8188&41.8 & 0.82&{\footnotesize Exams, Websites, Documents} &\XSolidBrush & \Checkmark& \XSolidBrush\\ \cmidrule(lr){1-9}
    ~\cite{wu-etal-2023-rethinking}&LEMON &22252 & 35.4&0.55 & {\footnotesize Daily Writing}& \Checkmark& \Checkmark& \XSolidBrush\\ \cmidrule(lr){1-9}
    ~\cite{hu-etal-2024-cscd}& CSCD-NS& 40000& 57.4&0.50 & Weibo & $\dagger$&\Checkmark &\Checkmark \\ \cmidrule(lr){1-9}
     ~\cite{10.1145/3616855.3635847}&AlipaySEQ &15522 & 7.4& 0.54& Search Engine Query& $\dagger$&\Checkmark &\Checkmark \\ 
    \bottomrule
    \end{tabular}
    \caption{Introduction to the existing CSC Dataset. ``TNS'' denotes the total number of sentences, while ``ASL'' refers to the average sentence length. ``AEPS'' indicates the average number of errors per sentence. The ``Source'' specifies the origin of the sentences. ``Real Error'' distinguishes whether the errors in the dataset are authentic or generated. ``Expert'' reflects the presence of manual annotations. ``Split'' indicates whether the dataset is divided into train and test (development) sets. Additionally, ``$\dagger$'' refers to dataset include pseudo-data generated through technical methods, to supplement the main manually annotated data (not calculated in the table).}
    \label{tab2}
\end{table*}
\subsection{Detector-Corrector Architecture}
The Detector-Corrector architecture (D-C) is a well-established framework for CSC, which follows a two-step process. First, the Detector identifies spelling errors in the input text, typically using Bi-GRU or a lightweight Transformer for binary classification. Second, the Corrector revises only the detected error locations, selecting the most probable character from the probability candidate set. However, the direct D-C architecture has its constraints, leading to ongoing research aimed at improving its efficiency.

Previous CSC methods are mainly based on the hard mask approach, which completely replaces detected erroneous characters, potentially discarding essential information of errors. To address these, Soft-Masked BERT~\cite{zhang-etal-2020-spelling} incorporates a Bi-GRU for error detection and introduces a soft masking mechanism that applies weighted masking based on error probability rather than direct replacement.

However, Soft-Masked BERT may diminish the visual and phonetic characteristics of errors during masking. To solve this, MDCSpell~\cite{zhu2022mdcspell} employs a late fusion strategy by directly using the original sentence as input to preserve the critical visual and phonetic features of typos. Furthermore, by integrating the hidden states of the detection and correction modules, it mitigates the misleading impact of typos on the context.

In addition, DR-CSC~\cite{huang-etal-2023-frustratingly} introduces an enhanced D-C architecture, dividing the process into three steps: detection, reasoning, and search. First, misspelled characters in the text are identified through a detection task. Next, the reasoning task examines the error category. Finally, the outcomes of the detection and reasoning are used to construct a search matrix based on the confusion set, optimizing the search range for candidate characters.

Furthermore, Bi-DCSpell~\cite{wu-etal-2024-bi} further optimizes D-C architecture. It uses a bidirectional interaction mechanism, enabling the detection and correction tasks to dynamically reinforce each other and employs a learnable control gate mechanism to regulate the intensity of information exchange between detection and correction modules.

\section{Exploration of CSC on LLMs}\label{sec4}
In the current era, where LLMs exert a profound influence on NLP research, there remains a relative scarcity of studies focusing on their application to CSC. Meanwhile, research based on PLMs continues to advance. This disparity arises because LLMs, as generative models, exhibit critical limitations in addressing CSC tasks, which will be discussed in detail in \textbf{Section~\ref{sec6.1}}. However, LLMs possess undeniable advantages over traditional CSC models in terms of domain adaptability and tolerance to diverse data~\cite{li2023effectiveness}. These attributes enable LLMs to perform context-sensitive adaptations rather than relying solely on rote memorization of fixed character pairs. Now, several studies try to address LLMs' issues in CSC, like over-correction or length change, and utilize LLMs' strengths.

\cite{liu-etal-2024-arm} proposes an innovative Alignment-and-Replacement Module (ARM). ARM employs an alignment method that incorporates a dynamic programming algorithm and Chinese character similarity calculations to closely align LLM outputs with the original sentences, minimizing discrepancies. To prevent the overcorrection of LLMs, a replacement strategy is applied. Notably, ARM requires no retraining or refine-tuning and can be directly integrated into existing PLMs based models.

\cite{li-etal-2024-c} introduces C-LLM, a CSC method based on character-level segmentation. It addresses the limitations of LLMs in handling character length constraints and phonetic similarity. By redesigning the character-level segmentation approach, conducting continuous pre-training to incorporate new vocabulary, and performing supervised fine-tuning, C-LLM simplifies the correction process to character copying and replacement and enhances the model's performance across general and multi-domain datasets.

\cite{zhou-etal-2024-simple} proposes a method that uses a training-independent and prompt-free framework. The approach employs a LLM purely as a language model and integrates it with a minimum distortion model to ensure semantic fidelity through phonetic and shape similarity. Additionally, length-reward and fidelity-reward strategies are incorporated to enhance generation fluency and minimize over-correction.
\section{Datasets and Evaluation Criteria}\label{sec5}
\subsection{Datasets of CSC}
CSC has long faced challenges related to limited data volume and low-quality datasets, which hinder training performance and evaluation accuracy. To tackle these issues, numerous researchers explore diverse methods to construct various datasets. A summary of existing datasets is provided in \textbf{Table~\ref{tab2}}, with detailed descriptions presented below.

\textbf{SIGHAN13/14/15}~\cite{wu2013chinese,yu2014overview,tseng2015introduction} are benchmarks for evaluating model performance. However, they present several significant issues: (1) These datasets consists of traditional Chinese characters, whereas the CSC task is conducted in a simplified Chinese characters. (2) All of them, especially SIGHAN13, contains numerous misuse of ``\begin{CJK}{UTF8}{gkai}的\end{CJK}'', ``\begin{CJK}{UTF8}{gkai}地\end{CJK}'', and ``\begin{CJK}{UTF8}{gkai}得\end{CJK}'', which are easily confused auxiliary words that modify adjectives, nouns, and verbs. (3) Since these datasets are from Chinese beginners, the sentences are relatively simple and contain numerous semantic ambiguities, making them unsuitable as benchmarks. \textbf{Wang271K}~\cite{wang-etal-2018-hybrid} is developed to address the scarcity of training data. The authors aimed to generate Chinese character pairs with visual and phonetic similarities. OCR and ASR are applied to recognize blurred Chinese characters and Mandarin speech corpus. Misrecognitions are interpreted as evidence of similarity between the incorrectly recognized and the corresponding correct characters. Then, these pairs are randomly inserted into sentences selected from \textit{The People's Daily} to complete the dataset. 

In addition, \textbf{ECSPell}~\cite{lv2023general} encompasses three domains: law, medicine, and official documentation. It draws data from judicial examination, medical network consultation, and official documents. During annotation, volunteers introduced single-character and phrase-level spelling errors reflective of real-world scenarios, followed by manual proofreading to minimize inconsistencies. \textbf{LEMON}~\cite{wu-etal-2023-rethinking} is a multi-domain dataset, encompassing seven fields: game, encyclopedia, contract, medical care, car, novel, and news. The errors are sourced from real-world daily writing and include both domain-specific terminology errors and mistakes that require contextual reasoning.

Furthermore, \textbf{MCSC}~\cite{jiang2022mcscset} is the first large-scale CSC dataset in the medical domain. Derived from Tencent's ``Yidia'' platform, the dataset is annotated by medical professionals and includes a medical confusion set, enabling the automatic generation of supplementary data. \textbf{CSCD-NS}~\cite{hu-etal-2024-cscd} represents the first CSC dataset specifically designed for native speakers, which is originated from authentic texts on Weibo. This dataset highlights key features of spelling errors made by native speakers, such as a high prevalence of homophone and word-level errors. Additionally, the research team introduced a pseudo-data generation method leveraging input methods, which produced 2 million high-quality pseudo-samples. \textbf{AlipaySEQ}~\cite{10.1145/3616855.3635847} is the first CSC dataset designed for real-world Chinese mobile search engine queries. It addresses spelling errors arising from the diversity of input methods and the heterogeneity of user groups. The dataset includes both manually annotated and automatically generated samples, characterized by short text length, diverse error types, and a distinctive distribution of high-frequency errors.

\subsection{Evaluation Criteria}
The current standard for evaluating CSC involves calculating precision ($P$), recall ($R$), and the F1-index ($F1$). These metrics are further categorized into character-level and sentence-level, as well as detection and correction levels. Now, the majority of studies employ $P$, $R$, and $F1$ calculations for both error detection and correction at the sentence level. The calculation method of $F1$ is as follows:
\begin{equation}
        F1=(2\cdot P\cdot R)/(P+R)
\end{equation}
$P$ represents the proportion of right detections or corrections among all sentences modified by the model, while $R$ denotes the proportion of right detections or corrections among all error-containing sentences. Detection is deemed right if the identified position is accurate, whereas correction requires both the position and the corresponding label to be right.
\section{Challenges and Opportunities}\label{sec6}
This section is segmented into three parts. The first part discusses key issues associated with PLMs and reviews relevant studies trying addressing these challenges. The second part examines the limitations of LLMs in CSC and proposes potential solutions. The third part highlights the shortcomings of existing datasets and explores future directions.
\subsection{Challenges and Opportunities in PLMs}
Currently, some studies have highlighted issues with PLM-based models, including over-correction, limited generalization ability, and ineffective correction of continuous errors. Various methods have been proposed to address these challenges; however, these issues have not been completely resolved, suggesting that future research can build upon existing work to further tackle these problems.

\textbf{(1) Overcorrection and Low Generalization Ability:} In the masked language model training paradigm, the model often substitutes the correct characters with similar ones, leading to overcorrection. This training approach also causes the model to memorize only the character pairs it has encountered during training, leading to an inability to handle novel situations and resulting in limited generalization capability. To address these issues, ~\cite{wang2019confusionset} proposes a model that integrates a pointer network with a copy mechanism, which can identify and copy correct characters directly from the input sentence. ~\cite{wu-etal-2023-rethinking} suggests that masking 20\% of non-error tokens randomly encourages the model to better utilize context and enhance generalization. ~\cite{wei-etal-2024-training} presents the prior knowledge guided teacher, using prior knowledge to produce soft labels alongside real ones to mitigate overfitting.

\textbf{(2) Consecutive Errors:} Consecutive errors introduce considerable noise, which disrupts semantic understanding and errors will affect each other, making the correction process more complicated. For this problem, Scholars conduct some research.~\cite{wang2021dynamic} introduces a dynamic connected networks, which incorporates an attention mechanism to model dependencies between adjacent characters, effectively addressing the incoherence caused by consecutive errors.~\cite{shulin2022craspell} propose enhancing model robustness to multi-error texts by incorporating contextual noise modeling, which employs bidirectional KL divergence to constrain the distributional similarity between the original and the generated noisy context.~\cite{li-etal-2022-improving-chinese} presents a straightforward and effective method employing a multiple-inference mechanism to address continuous errors.

\subsection{Challenges and Opportunities in LLMs}\label{sec6.1}
Although LLMs outperform PLMs based models on many tasks, their performance in CSC remains inferior. This performance gap stems from several key limitations, which can be categorized into three issues, as illustrated in \textbf{Table~\ref{tab4}}.

\textbf{(1) Inability to Control Sentence Length:} This is the most critical limitation, since CSC requires the input and output sentences to maintain the same lengths. However, as a generative model, LLMs always alter sentence length despite attempts to control the output through the prompt, which significantly impacts their performance on CSC. In sentence 1, the LLM replaced the correct word ``\begin{CJK}{UTF8}{gkai}原 (originally)\end{CJK}'' with synonym ``\begin{CJK}{UTF8}{gkai}原本 (initially)\end{CJK}'' and added ``\begin{CJK}{UTF8}{gkai}修改后的句子是：(Revised sentence:)\end{CJK}'' at the beginning, significantly altering the length of the sentence. 

\textbf{(2) Overcorrection:} Due to LLMs training methodology, LLMs frequently transform uncommon but correct expressions into more common alternatives. In sentence 2,  despite being error-free, LLMs unnecessarily revise ``\begin{CJK}{UTF8}{gkai}服务生 (server)\end{CJK}'' to the more frequently used synonym ``\begin{CJK}{UTF8}{gkai}服务员 (waiter)\end{CJK}''. In addition to common synonyms,  overcorrection also exists in proper nouns, punctuation, date formats, etc.

\textbf{(3) Limited Understanding of Phonetic Information:} Owing to LLMs' poor phonetic comprehension ability, LLMs often fail to replace Chinese characters with similar-pinyin alternatives, even though the primary source of errors are phonetic similar characters. In sentence 3, instead of changing ``\begin{CJK}{UTF8}{gkai}生 (birth)\end{CJK}'' to the phonetically similar character ``\begin{CJK}{UTF8}{gkai}升 (up)\end{CJK}'' the LLM replaced ``\begin{CJK}{UTF8}{gkai}学 (school)\end{CJK}'' with ``\begin{CJK}{UTF8}{gkai}日 (day)\end{CJK}'' transforming the ``\begin{CJK}{UTF8}{gkai}升学宴 (graduation party)\end{CJK}''  into a ``\begin{CJK}{UTF8}{gkai}生日宴 (birthday party)\end{CJK}'' and thereby altering the meaning of the sentence.

Because of these drawbacks, there are not many studies on the application of LLMs in CSC. Therefore, leveraging LLMs strengths and addressing their shortcomings is a promising avenue for future research. To address the those issue, future studies can explore more advanced alignment methods rather than just being limited to character similarity~\cite{liu-etal-2024-arm} and try to utilize methods like retrieval enhancement to provide additional characters' information.

Recently, the growing popularity of DeepSeek has garnered increased attention to large-model reasoning, which also holds significant potential for CSC. Reasoning enables the simultaneous analysis of various aspects of Chinese characters, including pinyin, strokes, semantics, and grammar. It can even leverage knowledge graphs to identify factual inaccuracies. Furthermore, reasoning generates multiple potential correction outcomes, ranks them based on probability and semantic coherence, and selects the optimal solution. Additionally, reasoning is capable of addressing errors in long sentences, identifying issues through global semantic analysis rather than relying solely on surface-level matching. These capabilities demonstrate the vast potential of reasoning in CSC, suggesting that future work should explore its application to this domain.
\begin{table}[!ht]
\renewcommand{\arraystretch}{1.35}
\small
\centering
\begin{tabular}{c|l}
\hline
\multicolumn{2}{l}{\textbf{Sentence 1:}}\\
\hline
Input:&\begin{CJK}{UTF8}{gkai}我们\textcolor{blue}{原}就是夫妻\end{CJK}\\
Translation:&\small{We were \textbf{originally} a couple.}\\
LLMs' Output:&\begin{CJK}{UTF8}{gkai}\textcolor{red}{修改后的句子:}我们\textcolor{red}{原本}就是夫妻。\end{CJK}\\
Translation:&\scriptsize{\textbf{Revised sentence:} We were \textbf{initially} a couple.}\\
\hline
\multicolumn{2}{l}{\textbf{Sentence 2:}}\\
\hline
Input:&\begin{CJK}{UTF8}{gkai}\small{那个服务\textcolor{blue}{生}是第一天上班。}\end{CJK}\\
Translation:&\small{That \textbf{server} was on his first day on the job.}\\
LLMs' Output:&\begin{CJK}{UTF8}{gkai}\small{那个服务\textcolor{red}{员}是第一天上班。
}\end{CJK}\\
Translation:&\small{That \textbf{waiter} was on his first day on the job.}\\
\hline
\multicolumn{2}{l}{\textbf{Sentence 3:}}\\
\hline
Input:&\begin{CJK}{UTF8}{gkai}领导儿子的\textcolor{blue}{生}学宴。\end{CJK}\\
Translation:&\small{\textbf{The graduation party} for the leader’s son.}\\
LLMs' Output:&\begin{CJK}{UTF8}{gkai}领导儿子的\textcolor{red}{生日}宴。\end{CJK}\\
Translation:&\small{\textbf{The Birthday party} for the leader’s son.}\\
\hline
\end{tabular}
\caption{Examples of shortcomings of employing LLMs (e.g., GPT\-4 and Llama) in CSC. Incorrect characters are highlighted in red, with their original characters in blue. Additionally, the English corresponding to the modified part is marked bold.}
\label{tab4}
\end{table}
\begin{table}[!ht]
\renewcommand{\arraystretch}{1.35}
\small
\centering
\begin{tabular}{c|l}
\hline
\multicolumn{2}{l}{\textbf{Sentence 1:}}\\
\hline
Input:&\begin{CJK}{UTF8}{gkai}感恩时刻有\textcolor{red}{你你}的每小时\end{CJK}\\
Label:&\begin{CJK}{UTF8}{gkai}感恩时刻有\textcolor{red}{你你}的每小时\end{CJK}\\
Source:&LEMON\\
Translation:&\small{Thank you for every hour of your time}\\
\hline
\multicolumn{2}{l}{\textbf{Sentence 2:}}\\
\hline
Input:&\begin{CJK}{UTF8}{gkai}\small{跟\textcolor{red}{他}在一起的时候，...，老天爷把\textcolor{red}{她}送给我...}\end{CJK}\\
Label:&\begin{CJK}{UTF8}{gkai}\small{跟\textcolor{red}{她}在一起的时候，...，老天爷把\textcolor{red}{她}送给我...}\end{CJK}\\
Source:&SIGHAN\\
Translation:&\small{When I was with him,..., God sent her to me...}\\
\hline
\multicolumn{2}{l}{\textbf{Sentence 3:}}\\
\hline
Input:&\begin{CJK}{UTF8}{gkai}服务生说\textcolor{red}{咬}30元。\end{CJK}\\
Label:&\begin{CJK}{UTF8}{gkai}服务生说\textcolor{red}{要}30元。\end{CJK}\\
Source:&Wang271K\\
Translation:&\small{The waiter said it would cost 30 yuan.}\\
\hline
\end{tabular}
\caption{Examples of problematic sentences in the dataset, where problematic Chinese characters are marked in red.}
\label{tab3}
\end{table}
\subsection{Challenges and Opportunities in Datasets}
Although the number of datasets is increasing, some of these datasets contain annotation errors to varying degrees. Beyond these errors, additional issues persist. The most prominent problems are summarized below, with specific examples provided in \textbf{Table~\ref{tab3}}:

\textbf{(1) Grammatical Errors:} Some sentences contain grammatical errors that cannot be corrected through only replacement operation. These errors often disrupt sentence semantics significantly, which in turn affects the model's understanding of sentences. In sentence 1, the repetition of two ``\begin{CJK}{UTF8}{gkai}你 (you)\end{CJK}'' creates a grammatical error that spelling correction alone cannot resolve.

\textbf{(2) Incomplete Contexts:} Some sentences lack sufficient contextual information, making it impossible to determine the correct label based solely on the sentence. In sentence 2, the context does not clarify whether the third person refers to ``\begin{CJK}{UTF8}{gkai}她 (her)\end{CJK}'' or ``\begin{CJK}{UTF8}{gkai}他 (him)\end{CJK}'', so the right label cannot be got.

\textbf{(3) Multiple Correct Labels:} Certain errors in sentences can have more than one valid correction. In Sentence 3, ``\begin{CJK}{UTF8}{gkai}收 (charge)\end{CJK}'' is also a right label because ``\begin{CJK}{UTF8}{gkai}要 (need)\end{CJK}'' and ``\begin{CJK}{UTF8}{gkai}收 (charge)\end{CJK}'' are both phonetically and morphologically similar to ``\begin{CJK}{UTF8}{gkai}咬 (bite)\end{CJK}'' and they are in the same confusion set.

These issues negatively impact training effect and evaluation accuracy. Consequently, future research should prioritize developing higher-quality datasets or improving existing ones through technological or manual annotation methods.
\section{Conclusion}\label{sec7}
As a subfield of NLP, Chinese spelling correction (CSC) has garnered significant research attention. In this survey, we systematically reviews the development of CSC, covering task characteristics, model architectures, datasets, and evaluation indicators. We also identify key issues in current research, including limitations in models and datasets and propose potential directions for future advancements. We hope this survey can provide valuable insights for researchers and further advance the field in the era of LLMs.
\appendix

\bibliographystyle{named}
\bibliography{ijcai25}

\begin{thebibliography}{}

\bibitem[\protect\citeauthoryear{Cheng \bgroup \em et al.\egroup }{2020}]{cheng2020spellgcn}
Xingyi Cheng, Weidi Xu, Kunlong Chen, Shaohua Jiang, Feng Wang, Taifeng Wang, Wei Chu, and Yuan Qi.
\newblock Spellgcn: Incorporating phonological and visual similarities into language models for chinese spelling check.
\newblock In {\em Proceedings of the 58th Annual Meeting of the Association for Computational Linguistics}, pages 871--881, 2020.

\bibitem[\protect\citeauthoryear{Guo \bgroup \em et al.\egroup }{2021}]{guo2021global}
Zhao Guo, Yuan Ni, Keqiang Wang, Wei Zhu, and Guotong Xie.
\newblock Global attention decoder for chinese spelling error correction.
\newblock In {\em Findings of the Association for Computational Linguistics: ACL-IJCNLP 2021}, pages 1419--1428, 2021.

\bibitem[\protect\citeauthoryear{Hong \bgroup \em et al.\egroup }{2019}]{hong-etal-2019-faspell}
Yuzhong Hong, Xianguo Yu, Neng He, Nan Liu, and Junhui Liu.
\newblock {FASP}ell: A fast, adaptable, simple, powerful {C}hinese spell checker based on {DAE}-decoder paradigm.
\newblock In {\em Proceedings of the 5th Workshop on Noisy User-generated Text (W-NUT 2019)}, pages 160--169, Hong Kong, China, November 2019. Association for Computational Linguistics.

\bibitem[\protect\citeauthoryear{Hu \bgroup \em et al.\egroup }{2024}]{hu-etal-2024-cscd}
Yong Hu, Fandong Meng, and Jie Zhou.
\newblock {CSCD}-{NS}: a {C}hinese spelling check dataset for native speakers.
\newblock In {\em Proceedings of the 62nd Annual Meeting of the Association for Computational Linguistics (Volume 1: Long Papers)}, pages 146--159, Bangkok, Thailand, August 2024.

\bibitem[\protect\citeauthoryear{Huang \bgroup \em et al.\egroup }{2021}]{huang2021phmospell}
Li~Huang, Junjie Li, Weiwei Jiang, Zhiyu Zhang, Minchuan Chen, Shaojun Wang, and Jing Xiao.
\newblock Phmospell: Phonological and morphological knowledge guided chinese spelling check.
\newblock In {\em Proceedings of the 59th Annual Meeting of the Association for Computational Linguistics and the 11th International Joint Conference on Natural Language Processing (Volume 1: Long Papers)}, pages 5958--5967, 2021.

\bibitem[\protect\citeauthoryear{Huang \bgroup \em et al.\egroup }{2023}]{huang-etal-2023-frustratingly}
Haojing Huang, Jingheng Ye, Qingyu Zhou, Yinghui Li, Yangning Li, Feng Zhou, and Hai-Tao Zheng.
\newblock A frustratingly easy plug-and-play detection-and-reasoning module for {C}hinese spelling check.
\newblock In {\em Findings of the Association for Computational Linguistics: EMNLP 2023}, pages 11514--11525, Singapore, December 2023.

\bibitem[\protect\citeauthoryear{Ji \bgroup \em et al.\egroup }{2021}]{ji-etal-2021-spellbert}
Tuo Ji, Hang Yan, and Xipeng Qiu.
\newblock {S}pell{BERT}: A lightweight pretrained model for {C}hinese spelling check.
\newblock In {\em Proceedings of the 2021 Conference on Empirical Methods in Natural Language Processing}, pages 3544--3551, Online and Punta Cana, Dominican Republic, November 2021.

\bibitem[\protect\citeauthoryear{Jiang \bgroup \em et al.\egroup }{2022}]{jiang2022mcscset}
Wangjie Jiang, Zhihao Ye, Zijing Ou, Ruihui Zhao, Jianguang Zheng, Yi~Liu, Bang Liu, Siheng Li, Yujiu Yang, and Yefeng Zheng.
\newblock Mcscset: A specialist-annotated dataset for medical-domain chinese spelling correction.
\newblock In {\em Proceedings of the 31st ACM international conference on information \& knowledge management}, pages 4084--4088, 2022.

\bibitem[\protect\citeauthoryear{Li \bgroup \em et al.\egroup }{2022a}]{li-etal-2022-wspeller}
Fangfang Li, Youran Shan, Junwen Duan, Xingliang Mao, and Minlie Huang.
\newblock {WS}peller: Robust word segmentation for enhancing {C}hinese spelling check.
\newblock In {\em Findings of the Association for Computational Linguistics: EMNLP 2022}, pages 1179--1188, Abu Dhabi, United Arab Emirates, December 2022.

\bibitem[\protect\citeauthoryear{Li \bgroup \em et al.\egroup }{2022b}]{li-etal-2022-improving-chinese}
Jiahao Li, Quan Wang, Zhendong Mao, Junbo Guo, Yanyan Yang, and Yongdong Zhang.
\newblock Improving {C}hinese spelling check by character pronunciation prediction: The effects of adaptivity and granularity.
\newblock In {\em Proceedings of the 2022 Conference on Empirical Methods in Natural Language Processing}, pages 4275--4286, Abu Dhabi, United Arab Emirates, December 2022.

\bibitem[\protect\citeauthoryear{Li \bgroup \em et al.\egroup }{2022c}]{li-etal-2022-learning-dictionary}
Yinghui Li, Shirong Ma, Qingyu Zhou, Zhongli Li, Li~Yangning, Shulin Huang, Ruiyang Liu, Chao Li, Yunbo Cao, and Haitao Zheng.
\newblock Learning from the dictionary: Heterogeneous knowledge guided fine-tuning for {C}hinese spell checking.
\newblock In {\em Findings of the Association for Computational Linguistics: EMNLP 2022}, pages 238--249, Abu Dhabi, United Arab Emirates, December 2022.

\bibitem[\protect\citeauthoryear{Li \bgroup \em et al.\egroup }{2023}]{li2023effectiveness}
Yinghui Li, Haojing Huang, Shirong Ma, Yong Jiang, Yangning Li, Feng Zhou, Hai-Tao Zheng, and Qingyu Zhou.
\newblock On the (in) effectiveness of large language models for chinese text correction.
\newblock {\em arXiv preprint arXiv:2307.09007}, 2023.

\bibitem[\protect\citeauthoryear{Li \bgroup \em et al.\egroup }{2024}]{li-etal-2024-c}
Kunting Li, Yong Hu, Liang He, Fandong Meng, and Jie Zhou.
\newblock {C}-{LLM}: Learn to check {C}hinese spelling errors character by character.
\newblock In {\em Proceedings of the 2024 Conference on Empirical Methods in Natural Language Processing}, pages 5944--5957, Miami, Florida, USA, November 2024.

\bibitem[\protect\citeauthoryear{Liang \bgroup \em et al.\egroup }{2023}]{liang-etal-2023-disentangled}
Zihong Liang, Xiaojun Quan, and Qifan Wang.
\newblock Disentangled phonetic representation for {C}hinese spelling correction.
\newblock In {\em Proceedings of the 61st Annual Meeting of the Association for Computational Linguistics (Volume 1: Long Papers)}, pages 13509--13521, Toronto, Canada, July 2023.

\bibitem[\protect\citeauthoryear{Lin \bgroup \em et al.\egroup }{2024}]{lin-etal-2024-modalities}
Yongliang Lin, Zhen Zhang, Mengting Hu, Yufei Sun, and Yuzhi Zhang.
\newblock Modalities should be appropriately leveraged: Uncertainty guidance for multimodal {C}hinese spelling correction.
\newblock In {\em Proceedings of the 2024 Joint International Conference on Computational Linguistics, Language Resources and Evaluation (LREC-COLING 2024)}, pages 11463--11474, Torino, Italia, May 2024. ELRA and ICCL.

\bibitem[\protect\citeauthoryear{Liu \bgroup \em et al.\egroup }{2010}]{liu2010visually}
Chao-Lin Liu, Min-Hua Lai, Yi-Hsuan Chuang, and Chia-Ying Lee.
\newblock Visually and phonologically similar characters in incorrect simplified chinese words.
\newblock In {\em Coling 2010: Posters}, pages 739--747, 2010.

\bibitem[\protect\citeauthoryear{Liu \bgroup \em et al.\egroup }{2021}]{liu2021plome}
Shulin Liu, Tao Yang, Tianchi Yue, Feng Zhang, and Di~Wang.
\newblock Plome: Pre-training with misspelled knowledge for chinese spelling correction.
\newblock In {\em Proceedings of the 59th Annual Meeting of the Association for Computational Linguistics and the 11th International Joint Conference on Natural Language Processing (Volume 1: Long Papers)}, pages 2991--3000, 2021.

\bibitem[\protect\citeauthoryear{Liu \bgroup \em et al.\egroup }{2024}]{liu-etal-2024-arm}
Changchun Liu, Kai Zhang, Junzhe Jiang, Zirui Liu, Hanqing Tao, Min Gao, and Enhong Chen.
\newblock {ARM}: An alignment-and-replacement module for {C}hinese spelling check based on {LLM}s.
\newblock In {\em Proceedings of the 2024 Conference on Empirical Methods in Natural Language Processing}, pages 10156--10168, Miami, Florida, USA, November 2024.

\bibitem[\protect\citeauthoryear{Lv \bgroup \em et al.\egroup }{2023}]{lv2023general}
Qi~Lv, Ziqiang Cao, Lei Geng, Chunhui Ai, Xu~Yan, and Guohong Fu.
\newblock General and domain-adaptive chinese spelling check with error-consistent pretraining.
\newblock {\em ACM Transactions on Asian and Low-Resource Language Information Processing}, 22(5):1--18, 2023.

\bibitem[\protect\citeauthoryear{Shulin \bgroup \em et al.\egroup }{2022}]{shulin2022craspell}
L~Shulin, S~Shengkang, Y~Tianchi, et~al.
\newblock Craspell: a contextual typo robust approach to improve chinese spelling correction.
\newblock In {\em Proceedings of the Association for Computational Linguistics}, 2022.

\bibitem[\protect\citeauthoryear{Tseng \bgroup \em et al.\egroup }{2015}]{tseng2015introduction}
Yuen-Hsien Tseng, Lung-Hao Lee, Li-Ping Chang, and Hsin-Hsi Chen.
\newblock Introduction to sighan 2015 bake-off for chinese spelling check.
\newblock In {\em Proceedings of the Eighth SIGHAN Workshop on Chinese Language Processing}, pages 32--37, 2015.

\bibitem[\protect\citeauthoryear{Wang \bgroup \em et al.\egroup }{2018}]{wang-etal-2018-hybrid}
Dingmin Wang, Yan Song, Jing Li, Jialong Han, and Haisong Zhang.
\newblock A hybrid approach to automatic corpus generation for {C}hinese spelling check.
\newblock In {\em Proceedings of the 2018 Conference on Empirical Methods in Natural Language Processing}, pages 2517--2527, Brussels, Belgium, October-November 2018.

\bibitem[\protect\citeauthoryear{Wang \bgroup \em et al.\egroup }{2019}]{wang2019confusionset}
Dingmin Wang, Yi~Tay, and Li~Zhong.
\newblock Confusionset-guided pointer networks for chinese spelling check.
\newblock In {\em Proceedings of the 57th Annual Meeting of the Association for Computational Linguistics}, pages 5780--5785, 2019.

\bibitem[\protect\citeauthoryear{Wang \bgroup \em et al.\egroup }{2021}]{wang2021dynamic}
Baoxin Wang, Wanxiang Che, Dayong Wu, Shijin Wang, Guoping Hu, and Ting Liu.
\newblock Dynamic connected networks for chinese spelling check.
\newblock In {\em Findings of the Association for Computational Linguistics: ACL-IJCNLP 2021}, pages 2437--2446, 2021.

\bibitem[\protect\citeauthoryear{Wang \bgroup \em et al.\egroup }{2024}]{10.1145/3616855.3635847}
Yue Wang, Zilong Zheng, Zecheng Tang, Juntao Li, Zhihui Liu, Kunlong Chen, Jinxiong Chang, Qishen Zhang, Zhongyi Liu, and Min Zhang.
\newblock Towards better chinese spelling check for search engines: A new dataset and strong baseline.
\newblock In {\em Proceedings of the 17th ACM International Conference on Web Search and Data Mining}, WSDM '24, page 769–778, New York, NY, USA, 2024. Association for Computing Machinery.

\bibitem[\protect\citeauthoryear{Wei \bgroup \em et al.\egroup }{2023}]{wei-etal-2023-ptcspell}
Xiao Wei, Jianbao Huang, Hang Yu, and Qian Liu.
\newblock {PTCS}pell: Pre-trained corrector based on character shape and {P}inyin for {C}hinese spelling correction.
\newblock In {\em Findings of the Association for Computational Linguistics: ACL 2023}, pages 6330--6343, Toronto, Canada, July 2023.

\bibitem[\protect\citeauthoryear{Wei \bgroup \em et al.\egroup }{2024}]{wei-etal-2024-training}
Chi Wei, Shaobin Huang, Rongsheng Li, Naiyu Yan, and Rui Wang.
\newblock Training a better {C}hinese spelling correction model via prior-knowledge guided teacher.
\newblock In {\em Findings of the Association for Computational Linguistics: ACL 2024}, pages 13578--13589, Bangkok, Thailand, August 2024.

\bibitem[\protect\citeauthoryear{Wu \bgroup \em et al.\egroup }{2013}]{wu2013chinese}
Shih-Hung Wu, Chao-Lin Liu, and Lung-Hao Lee.
\newblock Chinese spelling check evaluation at sighan bake-off 2013.
\newblock In {\em Proceedings of the Seventh SIGHAN Workshop on Chinese Language Processing}, pages 35--42, 2013.

\bibitem[\protect\citeauthoryear{Wu \bgroup \em et al.\egroup }{2023}]{wu-etal-2023-rethinking}
Hongqiu Wu, Shaohua Zhang, Yuchen Zhang, and Hai Zhao.
\newblock Rethinking masked language modeling for {C}hinese spelling correction.
\newblock In {\em Proceedings of the 61st Annual Meeting of the Association for Computational Linguistics (Volume 1: Long Papers)}, pages 10743--10756, Toronto, Canada, July 2023.

\bibitem[\protect\citeauthoryear{Wu \bgroup \em et al.\egroup }{2024}]{wu-etal-2024-bi}
Haiming Wu, Hanqing Zhang, Richeng Xuan, and Dawei Song.
\newblock Bi-{DCS}pell: A bi-directional detector-corrector interactive framework for {C}hinese spelling check.
\newblock In {\em Findings of the Association for Computational Linguistics: EMNLP 2024}, pages 3974--3984, Miami, Florida, USA, November 2024.

\bibitem[\protect\citeauthoryear{Xu \bgroup \em et al.\egroup }{2021}]{xu-etal-2021-read}
Heng-Da Xu, Zhongli Li, Qingyu Zhou, Chao Li, Zizhen Wang, Yunbo Cao, Heyan Huang, and Xian-Ling Mao.
\newblock Read, listen, and see: Leveraging multimodal information helps {C}hinese spell checking.
\newblock In {\em Findings of the Association for Computational Linguistics: ACL-IJCNLP 2021}, pages 716--728, Online, August 2021.

\bibitem[\protect\citeauthoryear{Yin \bgroup \em et al.\egroup }{2024}]{yin-etal-2024-error}
Xunjian Yin, Xinyu Hu, Jin Jiang, and Xiaojun Wan.
\newblock Error-robust retrieval for {C}hinese spelling check.
\newblock In {\em Proceedings of the 2024 Joint International Conference on Computational Linguistics, Language Resources and Evaluation (LREC-COLING 2024)}, pages 6257--6267, Torino, Italia, May 2024.

\bibitem[\protect\citeauthoryear{Yu \bgroup \em et al.\egroup }{2014}]{yu2014overview}
Liang-Chih Yu, Lung-Hao Lee, Yuen-Hsien Tseng, and Hsin-Hsi Chen.
\newblock Overview of sighan 2014 bake-off for chinese spelling check.
\newblock In {\em Proceedings of The Third CIPS-SIGHAN Joint Conference on Chinese Language Processing}, pages 126--132, 2014.

\bibitem[\protect\citeauthoryear{Zhang \bgroup \em et al.\egroup }{2020}]{zhang-etal-2020-spelling}
Shaohua Zhang, Haoran Huang, Jicong Liu, and Hang Li.
\newblock Spelling error correction with soft-masked {BERT}.
\newblock In {\em Proceedings of the 58th Annual Meeting of the Association for Computational Linguistics}, pages 882--890, Online, July 2020.

\bibitem[\protect\citeauthoryear{Zhang \bgroup \em et al.\egroup }{2021}]{zhang2021correcting}
Ruiqing Zhang, Chao Pang, Chuanqiang Zhang, Shuohuan Wang, Zhongjun He, Yu~Sun, Hua Wu, and Haifeng Wang.
\newblock Correcting chinese spelling errors with phonetic pre-training.
\newblock In {\em Findings of the Association for Computational Linguistics: ACL-IJCNLP 2021}, pages 2250--2261, 2021.

\bibitem[\protect\citeauthoryear{Zhou \bgroup \em et al.\egroup }{2024}]{zhou-etal-2024-simple}
Houquan Zhou, Zhenghua Li, Bo~Zhang, Chen Li, Shaopeng Lai, Ji~Zhang, Fei Huang, and Min Zhang.
\newblock A simple yet effective training-free prompt-free approach to {C}hinese spelling correction based on large language models.
\newblock In {\em Proceedings of the 2024 Conference on Empirical Methods in Natural Language Processing}, pages 17446--17467, Miami, Florida, USA, November 2024.

\bibitem[\protect\citeauthoryear{Zhu \bgroup \em et al.\egroup }{2022}]{zhu2022mdcspell}
Chenxi Zhu, Ziqiang Ying, Boyu Zhang, and Feng Mao.
\newblock Mdcspell: A multi-task detector-corrector framework for chinese spelling correction.
\newblock In {\em Findings of the Association for Computational Linguistics: ACL 2022}, pages 1244--1253, 2022.

\end{thebibliography}

\end{document}